# Spatiotemporal Transformer Attention Network for 3D Voxel Level Joint Segmentation and Motion Prediction in Point Cloud

Zhensong Wei, Xuewei Qi, Zhengwei Bai, Guoyuan Wu, Saswat Nayak, Peng Hao, Matthew Barth
Yongkang Liu and Kentaro Oguchi

*Abstract*—Environment perception including detection, classification, tracking, and motion prediction are key enablers for automated driving systems and intelligent transportation applications. Fueled by the advances in sensing technologies and machine learning techniques, LiDAR-based sensing systems have become a promising solution. The current challenges of this solution are how to effectively combine different perception tasks into a single backbone and how to efficiently learn the spatiotemporal features directly from point cloud sequences. In this research, we propose a novel spatiotemporal attention network based on a transformer self-attention mechanism for joint semantic segmentation and motion prediction within a point cloud at the voxel level. The network is trained to simultaneously outputs the voxel level class and predicted motion by learning directly from a sequence of point cloud datasets. The proposed backbone includes both a temporal attention module (TAM) and a spatial attention module (SAM) to learn and extract the complex spatiotemporal features. This approach has been evaluated with the nuScenes dataset, and promising performance has been achieved.

## I. INTRODUCTION

With the goal of taking human drivers out of the equation, automated driving systems (ADS) have the potential to improve safety performance at the individual vehicle level and enhance accessibility for people with limited unsupervised mobility (e.g., elderly or disable). This can be fulfilled by leveraging a more comprehensive suite of onboard sensors and by eliminating human-related behaviors that lead to accidents, e.g., driving under the influence (DUI), distraction, fatigue, and recklessness [1].

Representative sensing technologies for AVs include camera, LiDAR (Light Detection And Ranging), radar, and ultrasonic sensors. In particular, LiDAR outperforms the other type of sensors in terms of providing precise depth measurements of 3D surroundings and retrieving high-fidelity information about the underlying scenery [2].

The rapid advances in deep learning (DL) techniques unlock a plethora of opportunities for extracting features from exteroceptive sensor data (e.g., LiDAR point clouds) and performing end-to-end tasks in automated driving [3]. For example, YOLO [4] defines an early paradigm for fast object detection and classification. Further progress on one-stage anchor-based detection has been made to improve the real-time performance and accuracy [5], or to adapt similar ideas to LiDAR-based 3D object detection using point cloud data [6]. Following this direction, various types of object detection models have been designed. As a recent breakthrough in natural language processing (NLP), "Transformer" leverages the self-attention mechanisms to explore positional relationship and representation within a sequence [7], and has been successfully extended to the field of computer vision [8]. However, much fewer efforts have been focused on the application of transformer to LiDAR point clouds, not to mention multiple tasks (e.g., detection, classification, and motion prediction) on 3D objects.

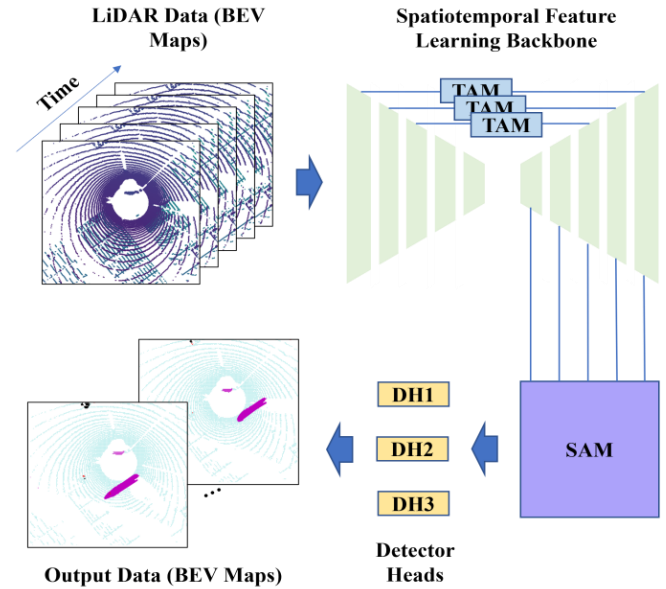

Figure 1. Joint semantic segmentation and motion prediction in point cloud with spatiotemporal attention.

In this research, we propose a novel spatiotemporal attention network based on transformer self-attention mechanism for joint semantic segmentation and motion prediction in point cloud at voxel level (see Figure 1). The designed spatiotemporal learning backbone includes both a temporal attention module (TAM) and a spatial attention module (SAM) to learn and extract the complex spatiotemporal features. The temporal attention module is designed to learn the motion feature inherited in a sequence of point clouds and the spatial attention module is designed to learn and extract the spatial features at different scales so that

Zhensong Wei, Zhengwei Bai, Guoyuan Wu, Saswat Nayak, Peng Hao and Matthew J. Barth are with the College of Engineering – Center for Environmental Research and Technology (CE-CERT), University of California, Riverside, CA 92507, USA. Email: zwei030@ucr.edu, zbai012@ucr.edu, gywu@cert.ucr.edu, snaya004@ucr.edu, haop@cert.ucr.edu, barth@ece.ucr.edu.

Xuewei Qi. Email: qixuewei@gmail.com.
Yongkang Liu and Kentaro Oguchi are with the Toyota Motor North America, R&D, InfoTech Labs. Email: yongkang.liu@toyota.com, kentaro.oguchi@toyota.com.

the semantic objects with different sizes can be accurately extracted and classified. The entire model is trained in an end-to-end manner with multiple task heads which are complementary to each other. The input of the model is a sequence of voxelized point clouds and the outputs of the model are voxel-level classification and motion predictions.

The rest of this paper is organized as follows. Section II presents the background information related to LiDAR-based tracking or/and trajectory prediction, semantic point cloud segmentation, and motion prediction, as well as transformer application. In Section III, the proposed spatiotemporal transformer pipeline is introduced, followed by the elaboration of validation results with the nuScenes dataset [39] in Section IV. The last section concludes this paper with further discussion on future steps.

## II. RELATED WORK

### A. Information Representation

Typical approaches on information representation for 2D or 3D environment perception in automated driving include, but are not limited to: a) bounding box identification [9]; b) anchor-free pointwise feature detection [10]; c) occupancy grid mapping (OGM) or voxelization [11]; and d) bird's eye view-based (BEV) mapping [12].

Bounding boxes or anchors provide a straightforward representation of 2D or 3D objects from the semantic or contextual perspective. Although being widely adopted, they fail to address the open-set scenarios where not all the objects have been well labeled as associated classes in the training dataset, not to mention those present in the complicated real-world situations. The same issue may apply to the anchor-free detection approaches. By contrast, cells or voxels are considered as building blocks for a generalized spatial representation of 2D or 3D environment. For example, OGM discretizes 3D point clouds into evenly spaced grids with binary variables indicating the presence of an object at that location (under the assumption that at least one point is occupied in the grid). To better correlate grid level information across time, BEV-based mapping extends OGM by discretizing the space around the ego-vehicle into cells and describing the associated class and dynamics of each of them. In the extreme cases, pixel-wise or point cloud-wise classification and prediction are needed for the applications with Unman Aerial Vehicles (UAVs).

### B. Semantic Point Cloud Segmentation

Semantic segmentation on 3D point clouds aims to classify each point or voxel independently into different homogeneous regions such that points or voxels inside the same region exhibit similar characteristics (e.g., same semantic meaning). This task is challenging mainly due to the complicated structure (such as sparsity and heterogeneity in sampling density) and inherent noises of point cloud data. Over the past few years, semantic point cloud segmentation with deep learning has been attracting more and more attention. Previous attempts can generally be classified into three paradigms: point-based [13], spatially discretized [14], and projection-based [15]. Point-based approaches take point clouds directly as the input without significant effort in data transformation, which may be further divided into point-wise multilayer perceptron (MLP) based, point convolution, recursive neural network (RNN) based, and graph-based [16]. By contrast, spatially discretized approaches separate 3D cloud points into different volumetric cells based on their spatial relationship. Although such data aggregation process is natural from the perspective of 3D data structure, it may result in the dilemma between resolution requirements and computational loads. To handle large-scale point cloud datasets, projection-based approaches project 3D point clouds into multiple 2D planes or apply spherical projection to represent 3D information into 2D images. Such dimension reduction techniques may facilitate the extended use of those state-of-the-art networks for 2D images, but they inevitably suffer from information loss (e.g., occlusion).

### C. Motion Prediction

Traditional work on motion prediction or trajectory prediction is performed at the object level based on the history information, which is considered as the downstream stage of object detection and tracking [17, 18]. Therefore, the prediction results heavily rely on the accuracy and reliability of trajectory acquisition from upstream stage(s). Recent studies on end-to-end joint tasking for 3D object detection, tracking as well as trajectory forecasting have shown promising results [19, 20], but they all depend on the bounding box identification with well-labeled training datasets and their applicability to automated driving systems (ADS) for real-world scenarios is questionable. As aforementioned, more generalized spatial information representation, such as voxel or occupancy grids, can alleviate this concern without the need to acquire bounding boxes for objects. In addition, multistep dynamic OGM [21] and BEV-based mapping [22] can better represent the temporal association between grids, which is suitable for the motion prediction purpose.

### D. Transformer Self-Attention

Achieving tremendous success in the domain of NLP [7], Transformers, designed for modeling long-range dependencies in the data, has gained increasing attention from researchers in the domain of computer vision (CV). Recently investigations demonstrated the promising results on certain CV tasks, such as image classification [8] and joint vision-language modeling [23].

Inspired by the power of the self-attention layer and structure of the Transformer, some works tried to replace partial or all spatial convolutional layers in some popular CV backbones, like ResNet [24-26]. Specifically, local-window-based self-attention for each pixel was applied and these models achieved slightly better performance with the trade-off of costly computational resources than the convolutional networks [24]. Another different way of investigation was to augment a standard convolutional layer with self-attention layers or Transformers. For instance, endowing the capability to model the dependencies or heterogeneous in the data, self-attention layers can augment the backbone networks [27-30] or the head networks [31, 32]. Recently, transformer-based vision backbone designs have been proposed and achieved outstanding results in CV tasks, such as the Vision

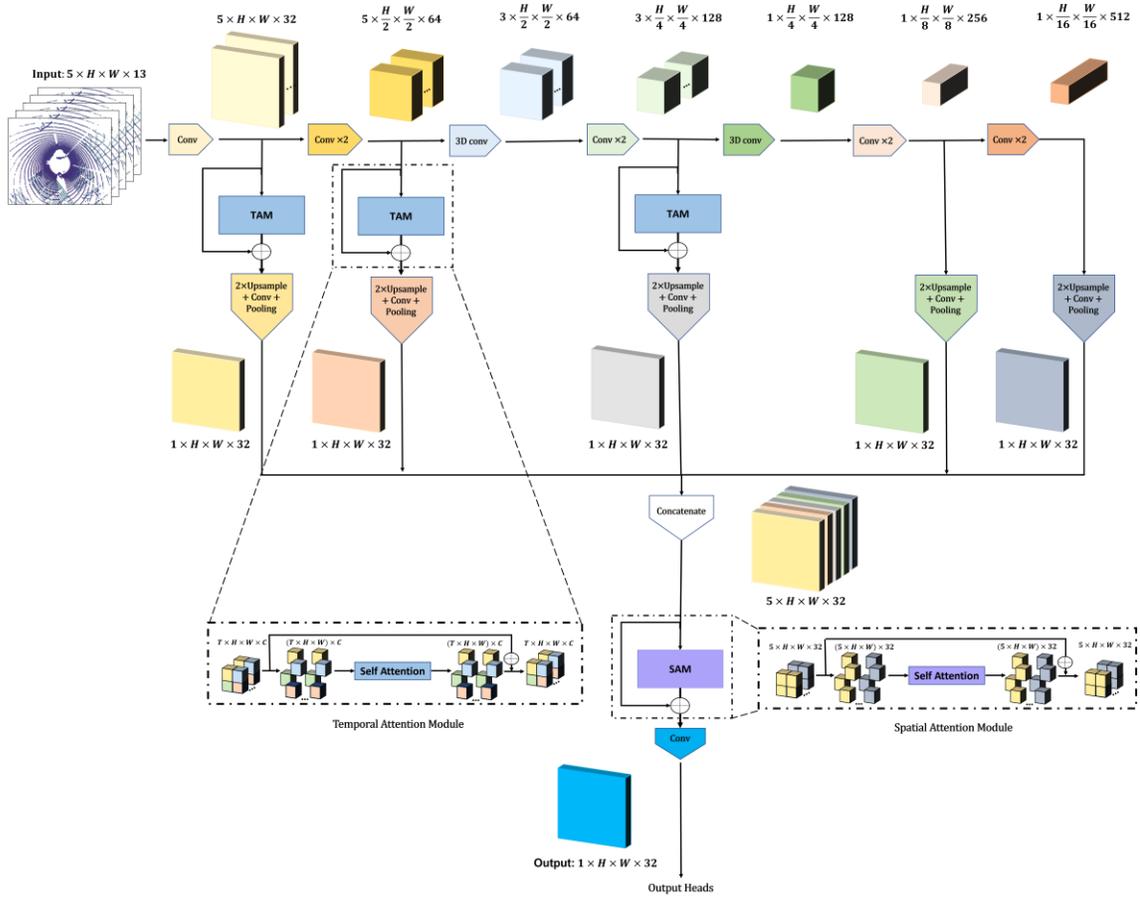

Figure 2. Spatiotemporal transformer network. Three temporal transformers are applied at 3 levels in the down-sampling path to extract the temporal features. One spatial transformer is applied after the concatenation of all levels in the down-sampling path to extract the spatial features among all dimensions.

Transformer (ViT) [33] and its follow-ups [34-36].

## III. METHODOLOGY

In this section, we present a proposed pipeline for roadside LiDAR-based joint perception and motion prediction for cooperative driving automation: (1) data representation from 3D LiDAR point clouds to BEV images; (2) Transformer-embedded spatiotemporal pyramid network as the backbone; and (3) task-specific heads for voxel-level classification and motion prediction.

### A. Input and Output Data Representation

**Input Representation**: Since 3D point clouds are usually collected in a sparse and nonuniform formation, we convert the points into voxels with a fixed resolution on the three dimensions, respectively. For the input, we consider five timesteps of points within 64 meters in X and Y dimensions, and five meters in Z dimensions, therefore the field of view (FOV) is of size $5 \times 64m \times 64m \times 5m$. The five time steps of data are sampled from a total of 20 past consecutive time frames with a frame skip equals to three to save computational power. Similar to [12], we use a binary state for each voxel stating whether it was occupied by at least one LiDAR point. With a resolution of $0.25m \times 0.25m \times 0.4m$, this result in a sequence of 13-channel binary pseudo-image of size $5 \times 256\ (H) \times 256\ (W) \times 13\ (C)$ cells.

Compared with other data representation methods such as point cloud based [13] or view-based network [37], voxel-based representation can utilize the well-established convolution and pooling operations for feature extraction, therefore making the computation much more efficient than other representations.

**Output Representation**: There are three output heads and each of them is organized differently. For the *motion prediction* head, we predict the future trajectory of each cell in the FOV with output represented as $\{X^t = (x_t, y_t)\}_{t=1}^{T}$, where $X^t \in R^{T \times H \times W \times 2}$ is the future position of the cell at each time step $t$; $x_t$ and $y_t$ are the predicted position in X and Y axis, respectively; and $T$=20 is the total number of predict frames. For the *cell classification* head, we predicted the class of each cell of the current time step with output $\in R^{H \times W \times C}$, where C= 5 is the total number of classes. For the *state estimation* head, we predict the probability of static for each cell at the current step with output $\in R^{H \times W}$. Note that all the three heads omit the height dimension since the objects are assumed to be moving on the ground without overlapping.

### B. Transformer-embedded Backbone

Transformers are based on attention mechanisms with input denoted as a sequence of discrete tokens $x \in R^{C \times d_x}$, where C represents the number of tokens in the sequence and

$d_x$ is the dimension of the feature vector in each token. A learnable positional encoding feature is also included to encode the positioning information of each token. Given such input, the queries (*Q*), keys (*K*), and values (*V*) are then calculated using projections as shown below:

$$Q = xW^Q, \quad K = xW^K, \quad V = xW^V \quad (1)$$

where $W^Q \in R^{d_x \times d_Q}, W^K \in R^{d_x \times d_K}$, and $W^V \in R^{d_x \times d_V}$ are parameter matrices. Then the attention is calculated using the dot products between *Q*, *K*, and *V*, as shown below:

$$\text{Attention}(Q, K, V) = softmax\left(\frac{QK^T}{\sqrt{d_K}}\right)V \quad (2)$$

Finally, the calculated attention will go through multilayer perceptron (MLP) to form the output *y*, which has the same size as input *x*.

$$y = \text{MLP}\big(\text{Attention}(Q, K, V)\big) + x \quad (3)$$

Since the input to the network is a time sequence of binary pseudo-image, our key idea is to utilize the well-established 2D and 3D convolutional layers as well as the self-attention mechanism of transformers to exploit the spatial and temporal features across different levels of the network. As shown in Figure 2, the input follows a horizontal down-sampling path with 2D convolutional blocks and 3D convolutional layers, and vertical transformer paths with convolutional layers, transformers and up-sampling layers. The 2D convolutional block consists of 2 convolutional layers that double the number of channels and decreases the size of the feature map by a factor of 4. The 3D convolutional layer decreases the dimension in the temporal channels and is appended after every 2D convolutional block until the temporal dimension is decreased to 1.

As aforementioned, the transformer is originally designed in the area of NLP and is not suitable for the input of images. Inspired by [38], we consider the intermediate feature maps of each down-sampling level to be a token set. As shown in Figure 2, both the temporal and spatial attention transformers require the input to be a two-dimensional token structure. Therefore, the feature maps in different layers in the backbone are first reshaped to the required dimension before feeding into the transformer. More specifically, assume that an input feature map of size $T_1 \times H \times W \times C$ is fed into the transformer. The tensor is first reshaped to the size of $(T_1 \times H \times W) \times C$ to match the token structure. Then, a learnable positional embedding feature of same size is added to the input feature map so that the tensor contains the spatiotemporal information during training. Finally, the output of the transformer is reshaped back to the size of input and is fed into the following convolutional layers. Since the same structure is applied multiple times at each down-sampling level, it is computationally expensive to operate on the original-sized feature maps at each level. Therefore, the feature map is down-sampled to H = W = 8 and is resized back to its original size before adding the input to form a residual connection.

The temporal attention module (TAM) takes the input sequence from each down-sampling block and learns the information embedded within the temporal dimension. The outputs are then upsampled and fed into a convolutional layer to form the same shape $1 \times H \times W \times 32$. The five outputs from each level of downsampling layers contain spatial information from different scales. Therefore, a spatial attention module (SAM) is used at the end to fuse the spatial information using the concatenated output from temporal transformer blocks. Note that the structures of the spatial and temporal transformer modules are identical. They are named differently due to the different physical meanings applied.

*C. Output Heads and Loss Function*

Three output heads are appended to the proposed backbone to obtain desired outputs: (1) motion prediction; (2) cell classification; and (3) state estimation, which consist of 2 convolutional layers with 1 batch normalization layer to formulate the dimensions of the output representation in each output head as mentioned above. To regulate the predicted motion and avoid small interference to static cells, such as background or stopped vehicles, the output of state estimation and cell classification heads are also utilized in the motion prediction head. More specifically, when a cell is classified as background from cell classification head or static from state estimation head, the motion prediction result for that cell will be set to 0. The cooperation of the three output heads can help avoid unnecessary computation and increase the training speed.

The loss function of the proposed spatiotemporal transformer network is adapted from the design of MotionNet [12], and is defined as the summation of six separate losses, which ca be globally optimized for the best performance:

$$\begin{aligned} L_{tot} &= L_{motion} + L_{cls} + L_{state} \\ &+ L_{Spatial} + L_{FTemporal} + L_{BTemporal} \end{aligned} \quad (4)$$

IV. EXPERIMENTS

In this section, we explain the experiment setup and compare the performance of the proposed network with the baseline using the nuScenes [39] dataset.

*A. Experiment Setup*

**Datasets.**

The nuScenes dataset is a public large-scale dataset for autonomous driving [39]. The dataset contains 1000 scenes collected from an entire onboard sensor suite, including cameras, LiDAR, radar, GPS, and IMU, with bounding box annotations on the keyframes. Among the 1000 scenes with a duration of 20s, 150 of the scenes are originally used for testing and do not contain any annotations. Therefore, LiDAR point clouds in the rest 850 scenes that are sampled at 20Hz are used for training (500), validation (100) and testing (250) in this study. To create the input for the network, keyframes are sampled at 2Hz in the training dataset and 1Hz in the validation and testing dataset. And a total of five consecutive frames including the keyframe sampled at 5Hz are combined in one clip as input. The lower sampling frequency in the validation and testing dataset is to avoid repetitive clips in the input. To create the target of the three output heads in the network, the annotated bounding boxes are preprocessed so that the voxelized cells are classified and the future positions are calculated. The original nuScenes dataset contains ground truth labels for 23 object classes. We define five classes for the classification task, including background, vehicle, pedestrian, bicycle and others. Note that the vehicle class in this study contains both car and bus from the original class in nuScenes dataset, and the "others" class contains all the other classes not included in the previous four defined classes. In total, there are 17,065 training clips, 1,719 validation clips, and 4,309 testing clips.

TABLE 1. Performance comparison between the baseline and the proposed spatiotemporal transformer network. The proposed network achieves better overall performance in the cell classification task and comparable performance in the motion prediction task.

| Method | Motion Prediction (m) | | | | | | Cell Classification (%) | | | | | | |
|---|---|---|---|---|---|---|---|---|---|---|---|---|---|
| | Static (speed = 0) | | Slow (speed <= 5m/s) | | Fast (speed > 5m/s) | | Background | Vehicle | Ped. | Bike | Others | **MCA** | **OA** |
| | Mean | Median | Mean | Median | Mean | Median | | | | | | | |
| MotionNet | 0.0230 | 0 | **0.2250** | **0.0957** | **0.8870** | **0.6120** | 95.6 | 90.5 | 79.1 | 15.1 | 71.4 | 70.3 | 94.5 |
| Ours | **0.0214** | 0 | 0.2426 | 0.0957 | 1.0504 | 0.7247 | **96.7** | 90.5 | 79.0 | **21.1** | 67.6 | **71.0** | **95.5** |

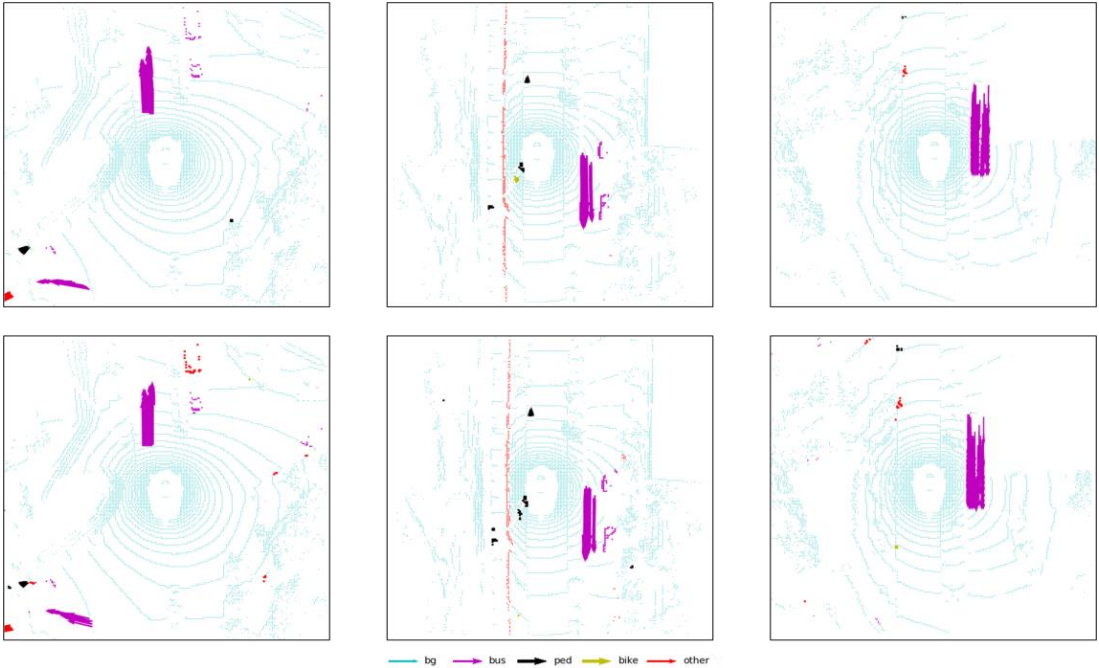

Figure 3. Sample motion prediction with cell classification network results. The arrow points to the predicted position of the cell with color representing its class. The predicted output (bottom row) shows accurate classification and precise motion prediction results compared to the ground truth (top row).

**Metrics and Training Environment.**

For the classification measurement, we use the following three performance metrics: (1) prediction accuracy for each class; (2) the overall classification accuracy (OA) for all cells; and (3) the mean category accuracy (MCA), which is the average accuracy of each class. For the motion prediction measurement, we divide the speed into three groups based on the speed range, which is static (speed = 0), slow (speed <= 5m/s), and fast (speed > 5m/s). The prediction error is calculated using the mean Euclidean distance of the predicted position between output and ground truth. Note that there are 20 time-steps (1s total) of future positions predicted in the output, and we only show the error of the last time step for simplicity.

The network is developed using a desktop computer with Nvidia GTX 2080 Ti GPU, a Core i9-9900K CPU running at 3.6 GHz, and 64 GB of RAM. The network is programmed using Python (version 3.7.0) with the PyTorch library (version 1.1.0).

*B. Comparison with SOTA*

The baseline used in this study is MotionNet [12], which has been proven to be effective in joint perception and motion prediction tasks. The backbone of MotionNet is a classic feature pyramid network, which consists of an up-sampling path and a down-sampling path, to extract multi-scale spatiotemporal features. The baseline and proposed model are trained in the same environment using the same dataset, and the result is shown in Table 1. As can be seen from the table, for the motion prediction task, the proposed method achieves better performance for static objects but performs slightly worse for moving objects. For the classification task, 95.5% overall accuracy and 71.0% mean category accuracy are achieved by the proposed method, both of which are higher than the baseline. Among all the listed classes, the cells labeled as background class are of the vast majority. The higher MCA shows that the transformer structure is capable of extracting the spatial features for the classification task. Note that the classification accuracy for the bike is small due to the rare occurrence of the class in the training dataset. The sample motion prediction with cell classification network results is shown in Figure 3. The predicted output shows accurate classification and precise motion prediction results compared to the ground truth. The inference time for the STAN is measured to be 0.154 sec, which is small enough for real-time performance.

TABLE 2. Performance comparison between different design of the network. The proposed network architecture achieves better overall performance in the cell classification task and motion prediction task.

| Method | Motion Prediction (m) | | | | | | Cell Classification (%) | | | | | | |
|---|---|---|---|---|---|---|---|---|---|---|---|---|---|
| | Static (speed = 0) | | Slow (speed <= 5m/s) | | Fast (speed > 5m/s) | | Background | Vehicle | Ped. | Bike | Others | **MCA** | **OA** |
| | Mean | Median | Mean | Median | Mean | Median | | | | | | | |
| STAN | **0.0220** | 0 | 0.283 | 0.0977 | **1.415** | 0.917 | 95.8 | 89 | 76.4 | 10.1 | 55.79 | 65.4 | **94.3** |
| TAN | 0.0224 | 0 | 0.271 | 0.0969 | 1.588 | 1.129 | 95.0 | 86.2 | 76.9 | 11.3 | 66.6 | 67.2 | 93.6 |
| SAN | 0.0230 | 0 | **0.263** | 0.0966 | 1.429 | 0.965 | 94.8 | 85.5 | 72.0 | 9.1 | 70.6 | 66.4 | 93.4 |
| RSTAN | 0.0299 | 0 | 0.287 | 0.0988 | 1.5990 | 1.2150 | 91.1 | 90.5 | 80.8 | 18.3 | 58.2 | **67.8** | 89.8 |

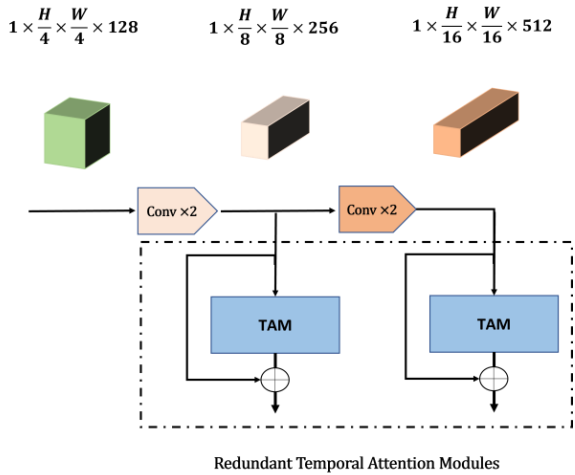

Figure 4. Redundant spatiotemporal transformer attention network, which appends temporal attention modules for all the vertical paths. The rest of the network structure is same as the one shown in Figure 2.

*C. Ablation Study*

To validate the design of the spatiotemporal transformer attention network (STAN), especially the effectiveness of the spatial and temporal attention modules, we compare our design with two variants: (1) Temporal transformer attention network (TAN) only, which removes the spatial attention module and keeps everything else same; (2) Spatial transformer attention network (SAN) only, which removes all the temporal attention modules and keeps everything else same; (3) Redundant spatiotemporal transformer attention network (RSTAN), which appends temporal attention modules for all the vertical paths even when the temporal dimension is already decreased to 1, as shown in Figure 4. To save training time, the networks are trained on a subset of the original dataset, which contains ~20% training data. The validation and testing set remain the same. As shown in Table 2, compared to the above three designs, STAN achieves the highest overall accuracy and smallest motion prediction error for static and fast-moving objects. The missing spatial attention module in the TAN makes the network harder to extract spatial information from different scales and causes an accuracy drop in the classification task. SAN performs the best in motion prediction tasks for slow objects, but performs worse than STAN in other metrics due to missing temporal attention modules. The redundant temporal attention modules in RSTAN make the performance overall the worst due to the unnecessary feature extractions in the additional temporal attention modules.

## V. Conclusions and Future Work

In this work, we proposed a novel spatiotemporal attention network based on a transformer self-attention mechanism for joint semantic segmentation and motion prediction in point cloud at the voxel level. The model has been extensively trained and evaluated on the open-source dataset. The results show that the designed model can efficiently learn and extract both spatial and temporal information inherited in the voxelized point cloud representation and achieve an overall classification accuracy of 95.5%. It can be applied to onboard applications as well as roadside sensing systems. Our future work includes more evaluations of the model on different types of point cloud datasets including data collected by roadside sensors.

## References

[1] Van Brummelena, J. et al. (2018). Autonomous Vehicle Perception: The Technology of Today and Tomorrow. *Transportation Research Part C*, 89, pp. 384–406
[2] Triess, L. et al. (2021). A Survey on Deep Domain Adaptation for LiDAR Perception. *arXiv:2106.02377v2*
[3] Grigorescu, S. et al. (2020). A Survey of Deep Learning Techniques for Autonomous Driving. *Journal of Field Robotics*, 37(3), pp. 362–386

[4] Redmon, J. et al. (2015). You Only Look Once: Unified, Real-Time Object Detection. *arXiv:1506.02640*
[5] Bochkovskiy, A. et al. (2020). YOLOv4: Optimal Speed and Accuracy of Object Detection. *arXiv:2004.10934*
[6] Simon, M. et al. (2018). Complex-YOLO: Real-time 3D Object Detection on Point Clouds. *arXiv:1803.06199*
[7] Vaswani, A. et al. (2017). Attention is All You Need. *In Advances in Neural Information Processing Systems*, pp. 5998–6008.
[8] Dosovitskiy, A. et al. (2021). An Image is Worth 16x16 Words: Transformers for Image Recognition at Scale. *In International Conference on Learning Representations*.
[9] Lang, A. et al. (2019). Pointpillars: Fast Encoders for Object Detection from Point Clouds. *In IEEE Conference on Computer Vision and Pattern Recognition*, pp. 12697–12705
[10] Duan, K. et al. (2019). CenterNet: Keypoint Triplets for Object Detection. *In IEEE Conference on Computer Vision and Pattern Recognition*.
[11] Hoermann, S. et al. (2018). Dynamic Occupancy Grid Prediction for Urban Autonomous Driving: A Deep Learning Approach with Fully Automatic Labeling. *In IEEE International Conference on Robotics and Automation (ICRA)*, pp. 2056–2063
[12] Wu, P. et al. (2020). MotionNet: Joint Perception and Motion Prediction for Autonomous Driving Based on Bird's Eye View Maps. *In IEEE Conference on Computer Vision and Pattern Recognition*.
[13] Qi, C. et al. (2017). PointNet: Deep Learning on Point Sets for 3D Classification and Segmentation. *In IEEE Conference on Computer Vision and Pattern Recognition*.
[14] Huang, J. and You, S. (2016). Point Cloud Labeling Using 3D Convolutional Neural Network. in *ICPR*.
[15] Lawin, F. et al. (2017). Deep Projective 3D Semantic Segmentation. In *CAIP*.
[16] Guo, Y. et al (2020). Deep Learning for 3D Point Clouds: A Survey. *IEEE Transactions on Pattern Analysis and Machine Intelligence*.
[17] Alahi, A. et al. (2016). Social LSTM: Human Trajectory Prediction in Crowded Spaces. In *Proceedings of the IEEE Conference on Computer Vision and Pattern Recognition*, pp. 961–971.
[18] Deo, N. and Trivedi, M. (2018). Convolutional Social Pooling for Vehicle Trajectory Prediction. In *Proceedings of the IEEE Conference on Computer Vision and Pattern Recognition Workshops*, pp. 1468–1476
[19] Luo, W. et al. (2018). Fast and Furious: Real Time End-to-end 3D Detection, Tracking and Motion Forecasting with a Single Convolutional Net. In *Proceedings of the IEEE conference on Computer Vision and Pattern Recognition*, pp. 3569–3577
[20] Weng, X. et al. (2021). PTP: Parallelized Tracking and Prediction with Graph Neural Networks and Diversity Sampling. IEEE Robotics and Automation Letters.
[21] Mohajerin, N. and Rohani, M. (2019). Multi-step Prediction of Occupancy Grid Maps with Recurrent Neural Networks. In *Proceedings of the IEEE Conference on Computer Vision and Pattern Recognition*, pp. 10600–10608
[22] Chen, X. et al. (2017). Multi-View 3D Object Detection Network for Autonomous Driving. In *Proceedings of the IEEE conference on Computer Vision and Pattern Recognition Workshop*.
[23] Alec Radford, Jong Wook Kim, Chris Hallacy, Aditya Ramesh, Gabriel Goh, Sandhini Agarwal, Girish Sastry, Amanda Askell, Pamela Mishkin, Jack Clark, Gretchen Krueger, and Ilya Sutskever. Learning transferable visual models from natural language supervision, 2021.
[24] Han Hu, Zheng Zhang, Zhenda Xie, and Stephen Lin. Local relation networks for image recognition. In Proceedings of the IEEE/CVF International Conference on Computer Vision (ICCV), pages 3464–3473, October 2019.
[25] Prajit Ramachandran, Niki Parmar, Ashish Vaswani, Irwan Bello, Anselm Levskaya, and Jon Shlens. Stand-alone selfattention in vision models. In Advances in Neural Information Processing Systems, volume 32. Curran Associates, Inc., 2019.
[26] Hengshuang Zhao, Jiaya Jia, and Vladlen Koltun. Exploring self-attention for image recognition. In Proceedings of the IEEE/CVF Conference on Computer Vision and Pattern Recognition, pages 10076–10085, 2020.
[27] Xiaolong Wang, Ross Girshick, Abhinav Gupta, and Kaiming He. Non-local neural networks. In IEEE Conference on Computer Vision and Pattern Recognition, CVPR 2018, 2018.
[28] Yue Cao, Jiarui Xu, Stephen Lin, Fangyun Wei, and Han Hu. Gcnet: Non-local networks meet squeeze-excitation networks and beyond. In Proceedings of the IEEE/CVF International Conference on Computer Vision (ICCV) Workshops, Oct 2019.
[29] Minghao Yin, Zhuliang Yao, Yue Cao, Xiu Li, Zheng Zhang, Stephen Lin, and Han Hu. Disentangled non-local neural networks. In Proceedings of the European conference on computer vision (ECCV), 2020.
[30] Jun Fu, Jing Liu, Haijie Tian, Yong Li, Yongjun Bao, Zhiwei Fang, and Hanqing Lu. Dual attention network for scene segmentation. In Proceedings of the IEEE Conference on Computer Vision and Pattern Recognition, pages 3146–3154, 2019.
[31] Han Hu, Jiayuan Gu, Zheng Zhang, Jifeng Dai, and Yichen Wei. Relation networks for object detection. In Proceedings of the IEEE Conference on Computer Vision and Pattern Recognition, pages 3588–3597, 2018.
[32] Jiayuan Gu, Han Hu, Liwei Wang, Yichen Wei, and Jifeng Dai. Learning region features for object detection. In Proceedings of the European Conference on Computer Vision (ECCV), 2018.
[33] Alexey Dosovitskiy, Lucas Beyer, Alexander Kolesnikov, Dirk Weissenborn, Xiaohua Zhai, Thomas Unterthiner, Mostafa Dehghani, Matthias Minderer, Georg Heigold, Sylvain Gelly, Jakob Uszkoreit, and Neil Houlsby. An image is worth 16x16 words: Transformers for image recognition at scale. In International Conference on Learning Representations, 2021.
[34] Li Yuan, Yunpeng Chen, Tao Wang, Weihao Yu, Yujun Shi, Francis EH Tay, Jiashi Feng, and Shuicheng Yan. Tokensto-token vit: Training vision transformers from scratch on imagenet. arXiv preprint arXiv:2101.11986, 2021.
[35] Xiangxiang Chu, Bo Zhang, Zhi Tian, Xiaolin Wei, and Huaxia Xia. Do we really need explicit position encodings for vision transformers? arXiv preprint arXiv:2102.10882, 2021.
[36] Kai Han, An Xiao, Enhua Wu, Jianyuan Guo, Chunjing Xu, and Yunhe Wang. Transformer in transformer. arXiv preprint arXiv:2103.00112, 2021.
[37] Su, H., Maji, S., Kalogerakis, E. and Learned-Miller, E., 2015. Multi-view convolutional neural networks for 3d shape recognition. In Proceedings of the IEEE international conference on computer vision (pp. 945-953).
[38] Prakash, Aditya, Kashyap Chitta, and Andreas Geiger. "Multi-modal fusion transformer for end-to-end autonomous driving." Proceedings of the IEEE/CVF Conference on Computer Vision and Pattern Recognition. 2021.
[39] Caesar et al. (2019). nuScenes: A Multimodal Dataset for Autonomous Driving. arXiv preprint arXiv:1903.11027, 2019.